\documentclass[10pt,twocolumn]{article}

\usepackage[a4paper,margin=18mm,top=18mm,bottom=20mm]{geometry}

\usepackage[T1]{fontenc}
\usepackage[utf8]{inputenc}
\usepackage{lmodern}
\usepackage{microtype}
\usepackage{times} 
\usepackage{float}
\usepackage{amsmath,amssymb,amsthm,bm}

\usepackage{tikz}
\usetikzlibrary{shapes.geometric, arrows.meta, positioning, decorations.pathreplacing, calligraphy}

\usepackage{graphicx}
\usepackage{booktabs}
\usepackage{multirow}
\usepackage{array}
\usepackage{caption}
\usepackage{natbib}
\usepackage{tabularx}
\usepackage{longtable}
\usepackage{tcolorbox}
\usepackage{enumitem}    
\usepackage{tcolorbox}   
\usepackage{amssymb}    
\usepackage[colorlinks=true,linkcolor=blue,citecolor=blue,urlcolor=blue]{hyperref}
\usepackage{xurl}
\usepackage{orcidlink} 

\usepackage{authblk}

\usepackage{titlesec}
\titlespacing*{\section}{0pt}{0.9em}{0.4em}
\titlespacing*{\subsection}{0pt}{0.7em}{0.3em}

\title{\bfseries Causal Analysis for Time Series Foundation Models}

\usepackage{authblk}
\usepackage{orcidlink}

\title{\bfseries Causal Analysis for Time Series Foundation Models}

\author[1,2,*]{Mathis Jander \orcidlink{0009-0004-9321-862X}}
\author[1]{Wouter van Heeswijk \orcidlink{0000-0002-5413-9660}}
\author[1]{Martijn Mes \orcidlink{0000-0001-9676-5259}}

\affil[1]{Faculty of Behavioral, Management and Social Sciences, University of Twente, Enschede, Netherlands}
\affil[2]{European Central Bank, Frankfurt am Main, Germany}
\affil[*]{Corresponding author: \href{mailto:mathis.jander@utwente.nl}{mathis.jander@utwente.nl}}

\date{}

\date{\today}

\begin{document}

\twocolumn[
\maketitle
\vspace{-6mm}

\vspace{2mm}
]

\begin{abstract}
\noindent
Transitioning from bespoke time series models towards time series foundation models changes the relationship of model and application from one-to-one to one-to-many. This shift introduces concentration risk as many, potentially high-risk, forecasting applications are exposed to the same biases and failure modes of a single time series foundation model. At the same time, this centralization allows for economies of scale in model development and validation. In this study we investigate how biases and failure modes of time series foundation models can be identified before deployment. We propose a causal analysis framework to investigate the ability of a time series foundation model to preserve time series patterns. To achieve this, we intervene on parameterized synthetic time series generators and measure the corresponding change in model output under \textit{ceteris paribus} conditions. We apply our causal analysis framework to \texttt{Chronos-2} and \texttt{TimesFM-2.5} and test them across six distinct time series patterns. We find safe configurations for trend and harmonic oscillation patterns. The results also indicate a bias in both models towards overestimating persistence, sudden failures for both models against the regime switch pattern and failure for \texttt{TimesFM-2.5} against the energy-release pattern. Our review of the original works for both models indicates that the findings might be explained by the data used for pretraining. We conclude our study with suggestions for further model development, recommendations for application-specific model selection, and a discussion of limitations and further research directions.
\end{abstract}

\section{Introduction}
Forecasts are widely used to inform decision-making in critical domains such as finance, energy and policy. At the same time, the time series models that we use to create these forecasts get less interpretable as the size of the datasets and number of trainable parameters increase. Traditionally, for each forecasting application a bespoke model was trained on an application-specific dataset, having model development and deployment within on team or organisation. With the creation of large, pretrained time series foundation models, such as \texttt{Chronos-2} \citep{ansariChronos2UniversalForecasting2025} and \texttt{TimesFM-2.5} \citep{dasDecoderOnlyFoundationModel2024}, the development and the use of time series models is separated further as the foundation models are developed by organizations, such as Amazon Science or Google Research, that then grant access to a larger developer community for many different downstream applications. While this shift from bespoke models for each application to foundation models is expected to create economies of scale in training and application \citep{ansariChronos2UniversalForecasting2025, dasDecoderOnlyFoundationModel2024, wooUnifiedTrainingUniversal2024}, it also creates two problems. First, it reduces the insight that the deploying organization has into model development and therefore what biases and failure modes it might have. Second, as one foundation model is intended to be used for many downstream applications, this introduces a systemic risk as the models biases and failure modes could affect many applications instead of being contained to one application as in the case of bespoke models. Looking at the economies of scale and the resulting risk profile, we argue that this centralization towards a few, widely used foundation models presents a necessity to scrutinize time series foundation models in more depth than bespoke models. Therefore, we will concern ourselves with the question of how we can identify biases and failure modes in time series foundation models before deploying them to any particular forecasting application.

This study makes four contributions to the current body of knowledge on time series foundation models. First, we define and apply a framework for causal analysis of time series foundation models. Second, we show dose-response relationships for \texttt{Chronos-2} and \texttt{TimesFM-2.5} for six distinct time series patterns that suggest a bias towards overestimating persistence and identify several failure modes. Third, through our analysis we find evidence for a bias towards overestimating persistence in both models and show sudden failures for the regime switch and energy-release generators. Fourth, based on our empirical findings we are able to inform model selection for downstream applications as well as the development of future time series foundation models.

The remainder of this study is structured as follows. We begin by reviewing the existing literature for answers on our research question in Section \ref{litrev}. We then define a causal analysis framework for time series foundation models and operationalize it with our experimental set-up in Section \ref{methodology}. We analyze the results in Section \ref{results} and compare our findings with prior studies in Section \ref{discussion}. In Section \ref{conclusion}, we conclude this study by summarizing our findings, acknowledging limitations and outlining avenues for further research. 

\section{Literature Review}
\label{litrev}

Before defining our causal analysis framework, we review the existing literature related to our research question. To understand whether there is a methodological gap in our ability to identify and potentially mitigate biases and failure modes of time series foundation models, we first review examples of how other engineering disciplines deal with technologies that possess a similar risk profile. Having these examples as a reference, we can compare literature on in time series foundation models against these standards. Hereby, we start with the original studies introducing the \texttt{Chronos-2} and the \texttt{TimesFM-2.5} models. We then widen our scope to other studies on causal analysis of time series foundation models. Finally, we will review adjacent work on explainability and robustness to further define the extent of the gap in literature.

While foundation models present novel challenges in machine learning, other engineering disciplines already developed practices to assess technologies that will be used in many, potentially high-risk, contexts. In drug development, new compounds are required to go through a three-stage process before being released to the market \citep{commissionerDrugDevelopmentProcess2020, AuthorisationMedicinesEuropean2009}. In the first stage, a new compound is tested \textit{in vitro} under \textit{ceteris paribus} conditions to assert high degree of causal control. The goal of this stage is to isolate causal effects of the compound. In the second stage, preclinical trials on animals allow for \textit{in vivo} testing of the compounds toxicity and other effects on an organism. Stage three then proceeds to \textit{in vivo} testing on humans to validate the expected treatment effect. After market release, a drug is continuously monitored for rare adverse effects that might have not been noticed during the previous stages. In automotive manufacturing, car development follows a similar process \citep{InternationalOrganizationStandardization2018, peterEssentialsNewCar, TestCenterComplete2025}. In a first stage components and subsystems are tested in dedicated test rigs that simulate temperature, abrasion and other causes compromising their integrity over time. These test rigs allow to isolate a factor such as temperature and observe its effect on a component under \textit{ceteris paribus} conditions. In a second stage, the full vehicle is tested under controlled conditions to validate the proper integration of all subsystems. This entails testing aerodynamics, fuel efficiency and crash behavior. After a car is released to market periodic inspections and defect investigations, potentially leading to recalls, are common depending on the jurisdiction. In both examples a new technology is first tested under \textit{ceteris paribus} conditions to isolate causal effects before then validating the technologies expected behavior under more realistic, although also more confounding, conditions.

In the respective studies introducing \texttt{Chronos-2} \citep{ansariChronos2UniversalForecasting2025} and \texttt{TimesFM-2.5} \citep{dasDecoderOnlyFoundationModel2024}, both time series foundation models are compared against other models by testing their predictive accuracy on benchmark datasets. For example, \texttt{TimesFM-2.5} outperforms every other model on the \texttt{Monash} benchmark, while \texttt{llmtime} is the second to last place. On the \texttt{Darts} benchmark on the other hand, \texttt{llmtime} is placed first and \texttt{TimesFM-2.5} third. We cannot discern with confidence whether these performance differences are a consequence of the the training data, model architecture or any other factor. We also cannot tell whether these findings will hold for future observations of the same data-generating process. As it is similar to \textit{in vivo} drug testing or test-driving a car, benchmarking alone does not allow us to understand what caused the observed performance results.

To the best of our knowledge, there is currently no literature applying causal analysis to study time series foundation models. While previous research explores causal inference from time series data, such as Granger causality \citep{grangerInvestigatingCausalRelations1969} or Pearlian structural causal models \citep{pearlIntroductionCausalInference2010a} to uncover causal relationships, these studies aim to infer cause and effect between variables from time series data and do not use time series data for model evaluation. Consequently, interventions on data-generating processes to study time series foundation model behavior remains unexplored. 

Besides the literature concerned with time series foundation models, there are several adjacent research streams within the larger machine learning field that pursue related aims. Research on explainable AI produced several techniques to quantify the relationship between inputs and outputs of machine learning models. Prominent techniques are SHAP \citep{lundbergUnifiedApproachInterpreting2017}, LIME \citep{ribeiroWhyShouldTrust2016}, and Partial Dependence Plots \citep{friedmanGreedyFunctionApproximation2001} which were developed for tabular data. The first two assign an importance score to each input feature of a given sample, based on idiosyncratic axioms and assumptions. Partial Dependence Plots on the other hand change a feature's value and observe the effect it has on the model output. Repeating this for a set of samples allows to estimate the average treatment effect of the intervention. Sweeping across several intervention values create a dose-response curve mapping the expected interaction between the input feature and the model output. While the intervention on realized samples is intuitive for most applications with tabular data, for time series this is not the case. Instead, we would like to intervene on patterns of the data-generating process itself, such as trend, seasonality or autocorrelation.

The robustness literature concerns itself with understanding how modifications of inputs degrades the performance of a machine learning model. This research stream can be roughly distinguished in two different subcategories \citep{braiekChapter3Machine2025}. First, robustness to corruption of input samples, either from noise \citep{hendrycksBenchmarkingNeuralNetwork2019} or intentional adversarial manipulation \citep{goodfellowExplaining2015}. Second, from shifts in the data distribution \citep{ rechtImageNetClassifiersGeneralize2019, kohWILDSBenchmarkIntheWild2021}. When we consider robustness analysis for our research question, we face three issues. First, because these methods modify observed values rather than controlling the underlying data-generating process, we face the same problem of meaning of an intervention on a time series as in the case of the explainable AI literature. Second, robustness analysis reveals to what extent model performance degrades under corruption of its inputs, but it does not reveal biases and failure modes of time series foundation models if we intervene on properties of the underlying data-generating process. While allowing us to analyze failure modes due to input corruption, robustness analysis does not allow us to identify biases and failure modes on uncorrupted samples. Third, the notion of robustness to data distribution shifts becomes unclear when we apply it to time series foundation models. Defining what the training data distribution is and what constitutes a distributional shift from that is not obvious considering the large and diverse amount of data time series foundation models are trained on \citep{ansariChronos2UniversalForecasting2025, dasDecoderOnlyFoundationModel2024}.

To conclude, other engineering disciplines established staged processes to test high-risk technologies such as pharmacology and cars. The process begins with  \textit{in vitro} causal analysis under \textit{ceteris paribus} conditions, trading off realism for the ability to identify causal mechanisms. In a second stage, the technology is then tested under more realistic conditions \textit{in vivo}. This two-stage process allows us to identify causal mechanisms and map dose-response relationships in isolation and then validate them under more realistic conditions. Reviewing the literature on time series foundation models suggests that current validation efforts are focused on the second stage through performance benchmarking experiments on realized observations of uncontrolled data-generating processes. We also find that related research efforts, namely explainable AI and robustness analysis, do not provide methodologies for causal analysis of time series foundation model's biases and failure modes. Together, these findings outline the methodological gap that we aim to address in the remainder of this study.

\section{Methodology}
\label{methodology}

In this section, we first propose our causal analysis framework for time series foundation models in Section \ref{subsec:causal_analysis_framework}. We then describe the experimental configuration for this study in Section \ref{subsec:exp_config}.

\subsection{Causal Analysis Framework}
\label{subsec:causal_analysis_framework} 

To formalize our causal analysis framework for time series foundation models, we define a mathematical notation for the data-generating process, the time series foundation model, and our logic for causal inference.

Let  $t\in \mathbb{N}$ represent a discrete point in time. A \textbf{generator} $G$ is a stochastic or deterministic function parameterized by a $d$-dimensional vector $\theta \in \Theta \subseteq \mathbb{R}^d$. A generator is a function that maps discrete points in time to time series values

\begin{equation}
    G: \mathbb{N} \rightarrow \mathbb{R}.
\end{equation}

A \textbf{trajectory} $\mathbf{y}$ is a single realization of length $T$ sampled from the generator
\begin{equation}
    \mathbf{y} = G_\theta(t) + \epsilon, \quad \mathbf{y} \in \mathbb{R}^T, \quad t \in \{1, \dots, T\}.
\end{equation}
By explicitly separating the structural parameters $\theta$ from the realization-specific noise $\epsilon$, we can isolate the impact of parameter interventions on model output.

A \textbf{time series foundation model} $M$ is a function that accepts a trajectory $\mathbf{y}$ as input and returns an output trajectory $M(\mathbf{y})$ of length $H$.
\begin{equation}
    M: \mathbb{R}^T \rightarrow \mathbb{R}^H.
\end{equation}

A \textbf{parameter statistic} $\delta$ is a function that compresses a trajectory into a scalar value
\begin{equation}
    \delta: \mathbb{R}^l \rightarrow \mathbb{R}, \quad l \in \{T, H \}.
\end{equation}
We apply $\delta$ to estimate a property related to a generator parameter $\theta_i$ from a realized trajectory.

\begin{figure}[htbp]
    \centering
    \begin{tikzpicture}[
        node distance=0.8cm and 0.35cm,
        >=Stealth,
        thick,
        circle_node/.style={draw, circle, minimum size=1.1cm, inner sep=1pt, align=center, font=\footnotesize, fill=white},
        rect_node/.style={draw, rectangle, minimum width=1.1cm, minimum height=0.6cm, inner sep=2pt, align=center, font=\footnotesize, fill=white},
        round_node/.style={draw, rectangle, rounded corners=3pt, minimum width=1.4cm, minimum height=0.6cm, inner sep=2pt, align=center, font=\footnotesize, fill=white},
        label_node/.style={align=center, font=\footnotesize\bfseries}
    ]

        \node[circle_node] (gen) {$G_\theta$};
        \node[rect_node, right=of gen] (in_traj) {$\mathbf{y}$};
        \node[circle_node, right=of in_traj, minimum size=0.8cm] (model) {$M$};
        \node[rect_node, right=of model] (out_traj) {$M(\mathbf{y})$};

        \node[round_node, below=of gen, yshift= 0.25cm] (interv) {$\operatorname{do}(\theta_i \!=\! \alpha)$};
        \node[round_node, below=of in_traj] (in_stat) {$\delta(\mathbf{y})$};
        \node[round_node, below=of out_traj] (out_stat){$\delta(M(\mathbf{y}))$};

        \draw[->] (interv) -- (gen);
        \draw[->] (gen) -- (in_traj);
        \draw[->] (in_traj) -- (model);
        \draw[->] (model) -- (out_traj);
        \draw[->] (in_traj) -- (in_stat);
        \draw[->] (out_traj) -- (out_stat);

        \draw[decorate, decoration={calligraphic brace, mirror, amplitude=4pt}, ultra thick] 
            ([yshift=-0.2cm]interv.south west) -- ([yshift=-0.2cm]interv.south east)
            node[pos=0.5, below=0.1cm, label_node] {Dose};

        \draw[decorate, decoration={calligraphic brace, mirror, amplitude=5pt}, ultra thick] 
            ([yshift=-0.2cm]in_stat.south west) -- ([yshift=-0.2cm]out_stat.south east)
            node[pos=0.5, below=0.1cm, label_node] {Response};

    \end{tikzpicture}
    \caption{The causal analysis framework. Parameter intervention on $\theta_i$ serves as controlled dose and parameter statistics on $\mathbf{y}$ and $M(\mathbf{y})$ serve as observed responses.}
    \label{fig:causal_analysis_sketch}
\end{figure}
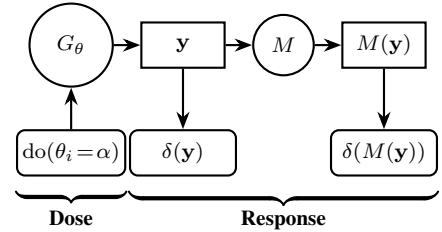

For causal analysis, we rely on Pearl's causal framework \citep{pearlIntroductionCausalInference2010a}. Under this framework we can define our components as a directed acyclic graph where $\theta \rightarrow \mathbf{y} \rightarrow M(\mathbf{y})$. A \textbf{parameter intervention} utilizes Pearl's $do$-operator to set our target parameter $\theta_i$ to a chosen value $\alpha$ while keeping all other parameters and the added noise sequence $\epsilon$ constant to create a \textit{ceteris paribus} condition. With this setup we can ensure that any observed change in $\delta(\mathbf{y})$ and $\delta(M(\mathbf{y}))$ is strictly caused by the change in our parameter $\theta_i$, eliminating potential confounding effects from other parameters or noise. With that, we then can establish a \textbf{dose-response relationship} by mapping the scalar responses $\delta(\mathbf{y})$ and $\delta(M(\mathbf{y}))$ across an interval of the intervened parameter $\theta_i$. Figure \ref{fig:causal_analysis_sketch} provides a sketch of the proposed causal analysis framework.

\subsection{Experimental Configuration}
\label{subsec:exp_config}

For the empirical study of biases and failure modes in time series foundation models, we apply our causal analysis framework to \texttt{Chronos-2} \citep{ansariChronos2UniversalForecasting2025} and \texttt{TimesFM-2.5} \citep{dasDecoderOnlyFoundationModel2024}, as they are frequently downloaded time series forecasting models on Hugging Face\footnote{\url{https://huggingface.co/models?pipeline_tag=time-series-forecasting&sort=downloads}}. We test both models across six different generators, each producing a distinct time series pattern. For all experimental setups, we fix  $T = 200$,  $H = 200$, and responses across $n = 50$ independent noise realizations. 

The first generator models a \textbf{random walk with drift}, where the trajectory evolves according to a standard stochastic drift process defined by
\begin{equation}
    y_t = y_{t-1} + \mu + \epsilon_t, \quad \epsilon_t \sim \mathcal{N}(0, \sigma^2),
\end{equation}
with the initial value fixed to $y_0 = 0.0$ and the noise scale to $\sigma = 1.0$. We perform a parameter intervention on the drift term, sweeping $\mu \in \{-0.05, -0.025, -0.005, 0.005, 0.025, 0.05\}$. As parameter statistic $\hat{\mu}$ we calculate the empirical mean of the first differences via 
\begin{equation}
   \hat{\mu} = \frac{1}{n-1}\sum_{t=2}^{n} (y_t - y_{t-1}). 
\end{equation}

To test how foundation models preserve temporal memory and persistence, we implement a \textbf{first-order autoregressive process}, AR(1), governed by
\begin{equation}
    y_t = c + \beta y_{t-1} + \epsilon_t, \quad \epsilon_t \sim \mathcal{N}(0, \sigma^2),
\end{equation}
with the fixed parameters $y_0 = 0.0$, $c = 0.0$, and $\sigma = 1.0$. We perform intervention on the autoregressive coefficient across the sweep $\beta \in \{-0.5, 0.0, 0.3, 0.6, 0.85, 0.98\}$. To measure responses, the parameter statistic $\hat{\beta}$ estimates the AR(1) $\beta$ coefficient. We implement the estimation of $\hat{\beta}$ with \texttt{statsmodels.tsa.ar\_model.AutoReg} \citep{seaboldStatsmodelsEconometricStatistical2010}.

For the third experiment, we produce a periodic time series via a \textbf{harmonic oscillator }, which generates a stationary sine wave and add noise such that
\begin{equation}
    y_t = A \sin\left(\frac{2\pi}{\lambda} t + \psi\right) + \epsilon_t, \quad \epsilon_t \sim \mathcal{N}(0, \sigma^2).
\end{equation}
Here, the amplitude is held at $A = 1.0$, the phase shift at $\psi = 0.0$, and the noise scale at $\sigma = 0.05$. We intervene directly on the wavelength parameter $\lambda \in \{5, 10, 25, 50, 75, 100\}$. The parameter statistic $\hat{\lambda}$ applies a Fast Fourier Transform (FFT) to the trajectory and extracts the most dominant frequency by peak height and converts it to wavelength. We implement $\hat{\lambda}$ with \texttt{scipy.signal.welch} \citep{virtanenSciPyFundamentalAlgorithms2020}.

To test whether structural breaks are preserved by both time series foundation models, we use a \textbf{regime switch generator with fixed dwell time} that features a piecewise linear trend with changing signs:
\begin{equation}
    y_t = y_{t-1} + m_t + \epsilon_t, \quad \epsilon_t \sim \mathcal{N}(0, \sigma^2).
\end{equation}
The drift alternates between $m_t \in \{1.0, -1.0\}$, reversing its state exactly every $\tau_{dw}$ steps under a noise scale of $\sigma = 0.05$. We intervene on the dwell time parameter $\tau \in \{5, 10, 25, 50, 75, 100\}$. The estimated dwell time $\hat{\tau}$ serves as the parameter statistic. We use \texttt{ruptures.Pelt} \citep{truongSelectiveReviewOffline2020} to calculate $\hat{\tau}$.

To capture nonlinear trigger events and critical thresholds, we use an \textbf{energy-release generator}. 
This generator is characterized by a build-up phase and a reset once a threshold $\kappa$ is crossed.
\begin{equation}
    y_t = y_{t-1} + s_t + |\epsilon_t|
\end{equation}
where $s_t = 0.2$  and $\epsilon_t \sim \mathcal{N}(0, \sigma^2)$ with $\sigma = 0.05$. Whenever $y_t > \kappa$, we reset $y_t = 0$. We intervene on the threshold $\kappa \in \{5, 10, 25, 50, 75, 100\}$. We calculate the parameter statistic $\hat{\kappa}$ by averaging peak values in our trajectory. We identify a peak if a time step is followed by a negative increment that is larger than half of the maximum value in the trajectory. 

For the last experiment, we test how well the time series foundation models preserve long-range dependence. We model \textbf{fractional Brownian motion}, fBm, process. The persistence of a trend, so an upward movement followed by an upward movement or a downward movement followed by a downward movement, is determined by the Hurst exponent $H$. For $H = 0.5$, the process equals a random walk. For $H > 0.5$ trends are persistent and for $H < 0.5$ antipersistent. We intervene on the Hurst exponent, sweeping $H \in \{0.15, 0.3, 0.45, 0.55, 0.7, 0.85\}$. As parameter statistic, we use $\hat{H}$ as the estimated Hurst exponent. To implement the generator we use \texttt{fbm.FBM}\footnote{\url{https://pypi.org/project/fbm/}} and to calculate $\hat{H}$ we use \texttt{hurst.compute\_Hc}\footnote{\url{https://pypi.org/project/hurst/}}. We summarize all six experiments with their respective generator, intervention sweep and parameter statistic in Table \ref{tab:experiments_summary}.

\begin{table}[htbp]
\centering
\caption{Summary of experiments with generators, intervention sweeps and parameter statistics.}
\label{tab:experiments_summary}
\footnotesize
\setlength{\tabcolsep}{3pt} 
\begin{tabular}{lllc}
\toprule
\textbf{Experiment} & \textbf{Generator} & \textbf{Intervention Sweep} & \textbf{$\delta$} \\
\midrule
1 & Random Walk & $\mu \in \{-0.05, -0.025, \dots, 0.05\}$ & $\hat{\mu}$ \\
2 & AR(1) & $\beta \in \{-0.5, 0.0, \dots, 0.98\}$ & $\hat{\beta}$ \\
3 & Harmonic Oscillator & $\lambda \in \{5, 10, 25, 50, 75, 100\}$ & $\hat{\lambda}$ \\
4 & Regime Switch & $\tau \in \{5, 10, 25, 50, 75, 100\}$ & $\hat{\tau}$ \\
5 & Energy-release & $\kappa \in \{5, 10, 25, 50, 75, 100\}$ & $\hat{\kappa}$ \\
6 & fBM & $H \in \{0.15, 0.3, \dots, 0.85\}$ & $\hat{H}$ \\
\bottomrule
\end{tabular}
\end{table}

\section{Results}
\label{results}

In this section, we review the results of applying our causal analysis framework to the experiments outlined in the previous section. For each experiment, we compare different intervention trajectories $\mathbf{y}$ with the same noise realization $\epsilon$  against the model outputs $M(\mathbf{y})$ of \texttt{Chronos-2} and \texttt{TimesFM-2.5}. We further compare the distributions of the trajectory parameter statistic $\delta(\mathbf{y})$ and  the model output parameter statistic $\delta(M(\mathbf{y}))$ for a given intervention $\operatorname{do}(\theta_i \!=\! \alpha)$ across all $n = 50$ noise realizations, as well as plot $\delta(\mathbf{y})$ against $\delta(M(\mathbf{y}))$ for a given intervention and noise realization. Taken together, we are able to characterize potential biases and failure modes of \texttt{Chronos-2} and \texttt{TimesFM-2.5} from the data generated by our experiments.

\begin{figure}[htbp]
    \centering
    \includegraphics[width=1\linewidth]{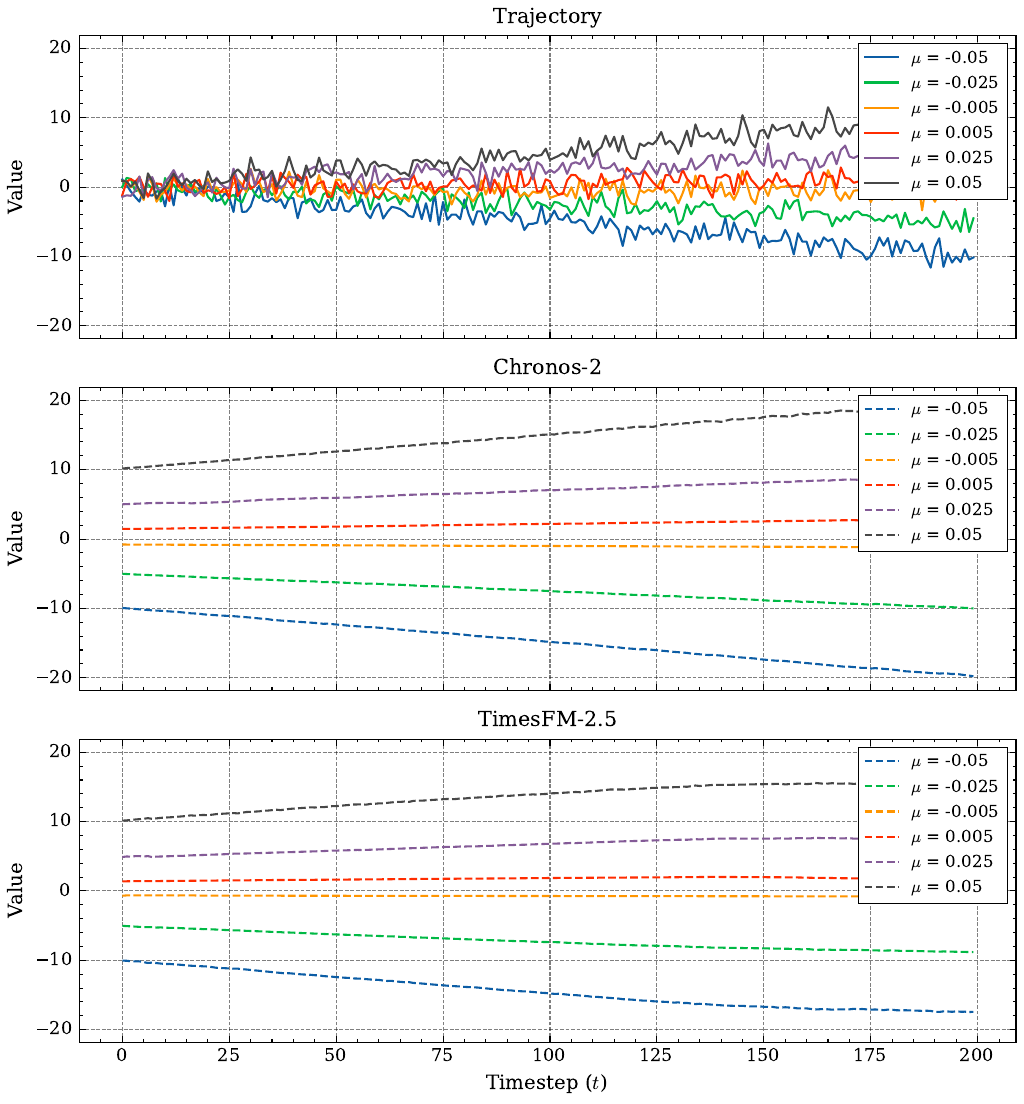}
    \caption{Trajectories across intervention sweep for a single noise realization in Experiment 1. Both time series foundation models seem to preserve the drift while removing noise. Towards the end of the trajectory, \texttt{TimesFM-2.5} seems to pull values towards zero.}
    \label{fig:exp1_traj}
\end{figure}

\begin{figure}[htbp]
    \centering
    \includegraphics[width=\columnwidth]{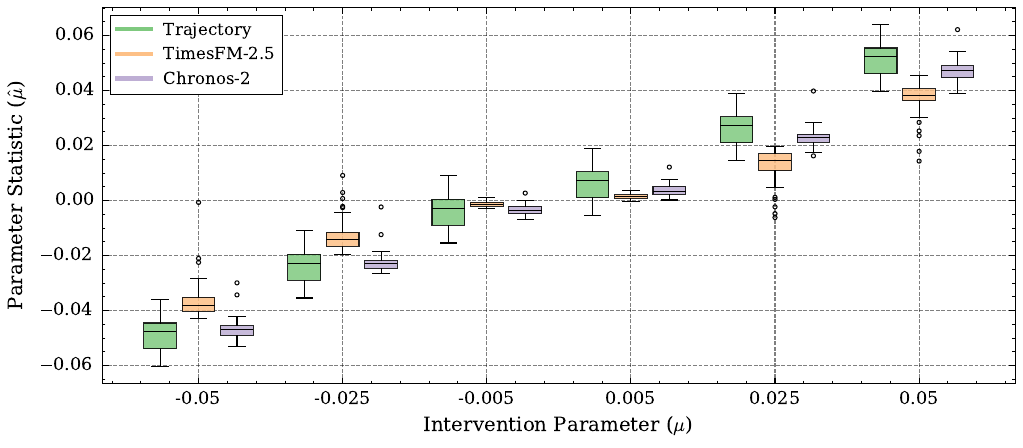}
    \caption{Distributions of parameter statistics for trajectory and model outputs across noise realizations in Experiment 1 ($n=50$). \texttt{TimesFM-2.5} displays lower magnitudes for drift than the trajectory.}
    \label{fig:exp1_box}
\end{figure}

\begin{figure}[htbp]
    \centering
    \includegraphics[width=1\linewidth]{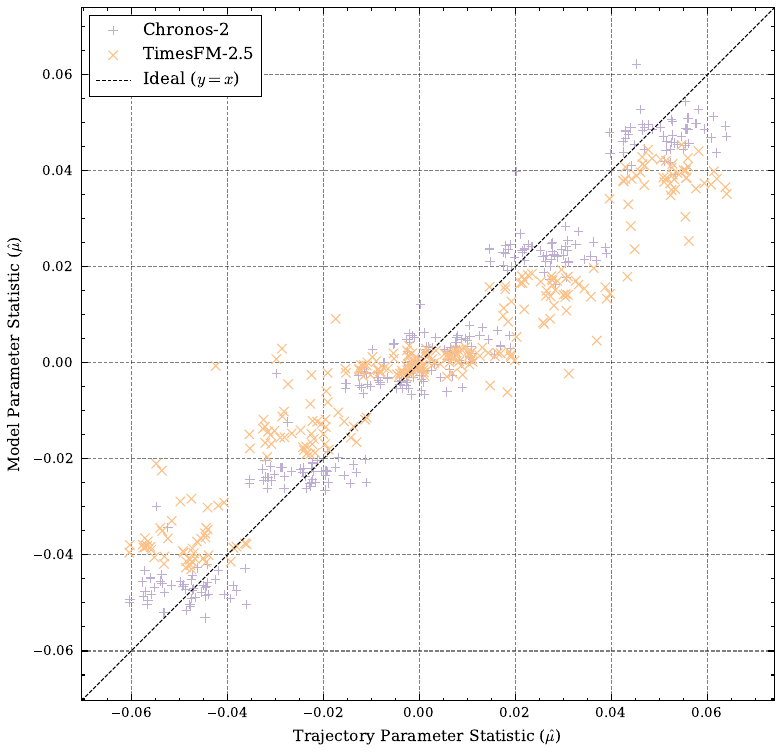}
    \caption{Comparison of model parameter statistic $\delta(M(\mathbf{y}))$ against trajectory parameter statistic $\delta(\mathbf{y})$ for Experiment 1 ($n = 6 \times 50 = 300$ per model). Both models center around the idea line, while \texttt{TimesFM-2.5} shows bias towards zero.}
    \label{fig:exp1_scatter}
\end{figure}

For \textbf{Experiment 1}, we find that both time series foundation models seem to smooth out the noise of the random walk process in their outputs, as shown in Figure \ref{fig:exp1_traj}. Figure \ref{fig:exp1_box} and \ref{fig:exp1_scatter} reveal that \texttt{Chronos-2}'s parameter statistic distribution is more closely aligned with the distribution of $\delta(\mathbf{y})$. Figure \ref{fig:exp1_box} also shows that \texttt{TimesFM-2.5} underestimates the magnitude of the drift parameter consistently for positive and negative values compared to \texttt{Chronos-2}. For the random walk with drift generator and the given range of interventions and noise realizations, \texttt{Chronos-2} appears better at preserving the drift of an input time series than \texttt{TimesFM-2.5}.

\begin{figure}[htbp]
    \centering
    \includegraphics[width=1\linewidth]{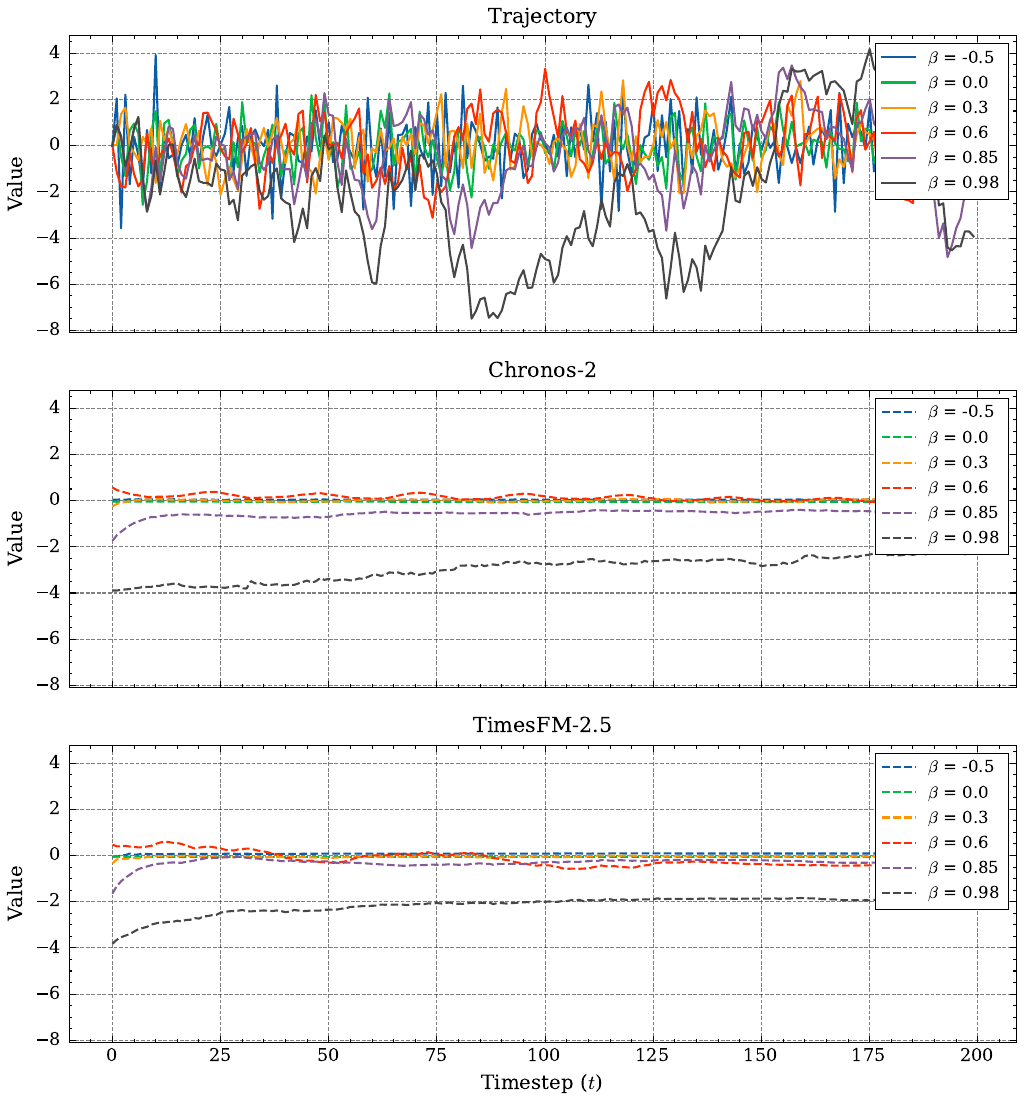}
    \caption{Trajectories across intervention sweep for a single noise realization in Experiment 2. Trajectories collapse close to flat lines for both time series foundation models.}
    \label{fig:exp2_traj}
\end{figure}

\begin{figure}[htbp]
    \centering
    \includegraphics[width=1\linewidth]{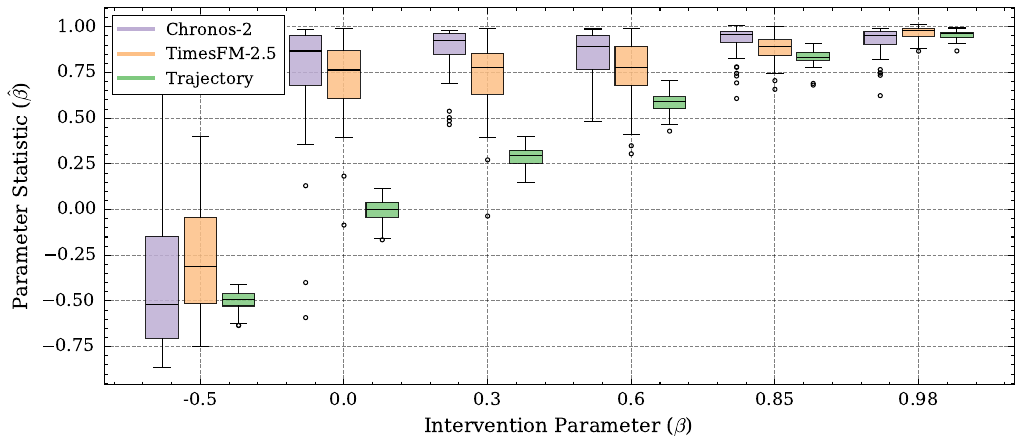}
    \caption{Distributions of parameter statistics for trajectory and model outputs across noise realizations in Experiment 2 ($n=50$). Both time series foundation models overestimate $\beta$ for lower values, except for $\operatorname{do}(\beta\!=\! -0.5)$, where \texttt{Chronos-2} seems to estimate $\beta$ better than \texttt{TimesFM-2.5}}
    \label{fig:exp2_box}
\end{figure}

\begin{figure}[htbp]
    \centering
    \includegraphics[width=1\linewidth]{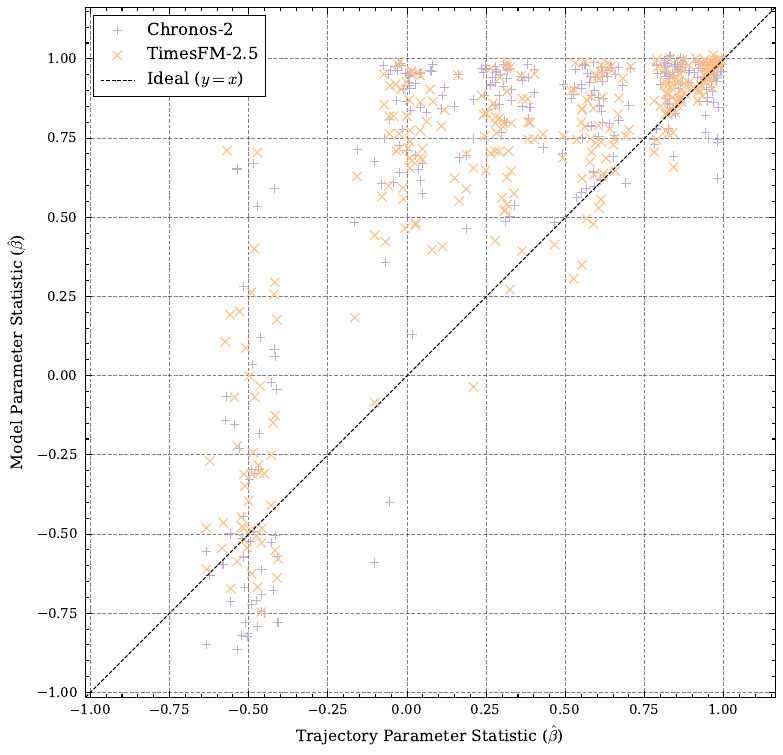}
    \caption{Comparison of model parameter statistic $\delta(M(\mathbf{y}))$ against trajectory parameter statistic $\delta(\mathbf{y})$ for Experiment 2 ($n = 6 \times 50 = 300$ per model). Both time series foundation models seem to produce outputs with a higher $\hat{\beta}$ than in the input trajectory for $0 < \beta < 0.85$.}
    \label{fig:exp2_scatter}
\end{figure}

The trajectory plot in Figure \ref{fig:exp2_traj} shows that both models' output flatlines the AR(1)-process in \textbf{Experiment 2} across all interventions of the displayed noise realization. For all interventions except $\operatorname{do}(\beta \!=\! 0.98)$, both models' output is close to zero. Figure \ref{fig:exp2_box} shows that both models produce high $\hat{\beta}$ values, even for low $\beta$ values. As $\beta$ values increase, the differences between the distribution of $\delta(\mathbf{y})$ and the respective distributions of $\delta(M(\mathbf{y}))$ of \texttt{Chronos-2} and \texttt{TimesFM-2.5} decrease. Figure \ref{fig:exp2_box} and \ref{fig:exp2_scatter} indicate that both models are biased towards consistently overestimating autocorrelation of lag one within our intervention range.

\begin{figure}[htbp]
    \centering
    \includegraphics[width=1\linewidth]{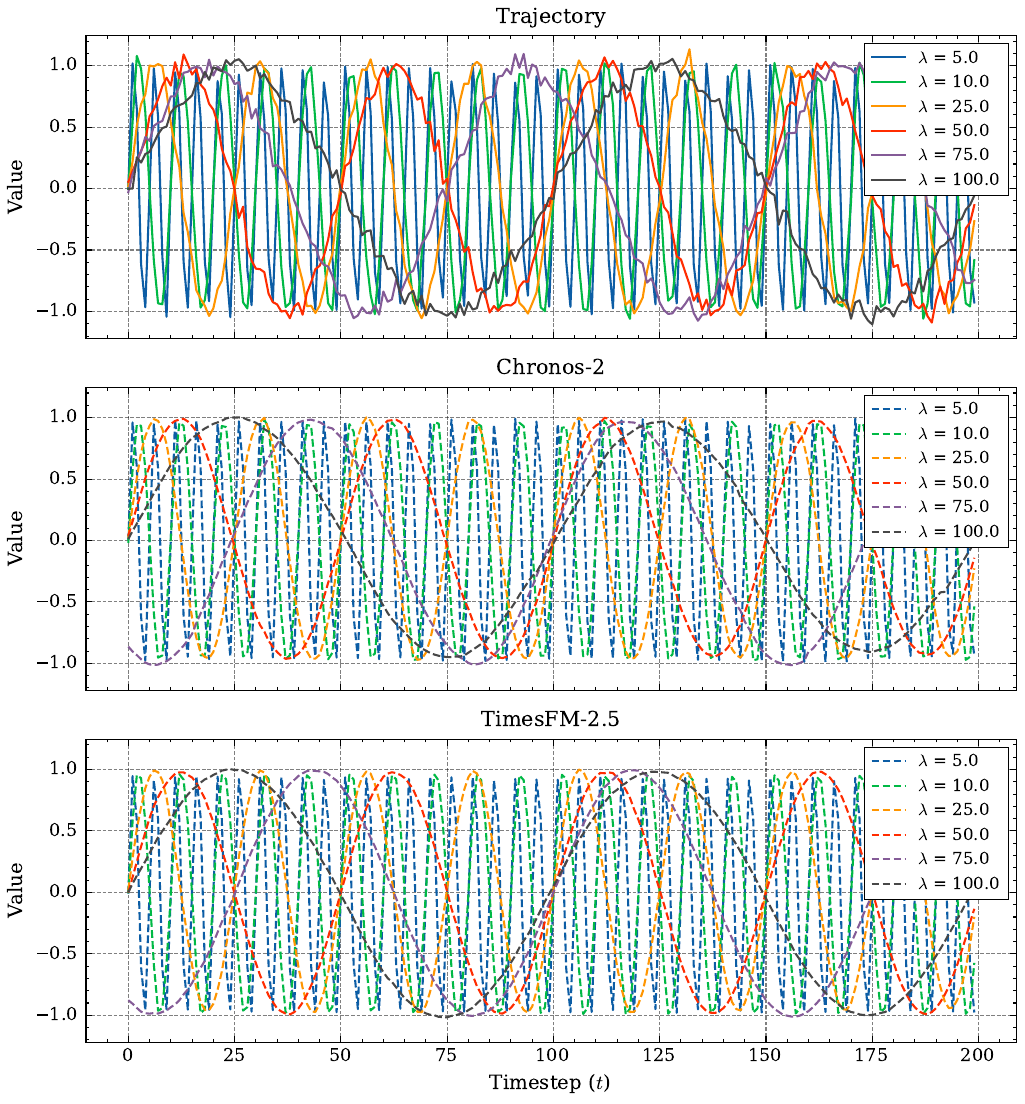}
    \caption{Trajectories across intervention sweep for a single noise realization in Experiment 3. Both time series foundation models seem to preserve wavelength across interventions.}
    \label{fig:exp3_traj}
\end{figure}

\begin{figure}[htbp]
    \centering
    \includegraphics[width=1\linewidth]{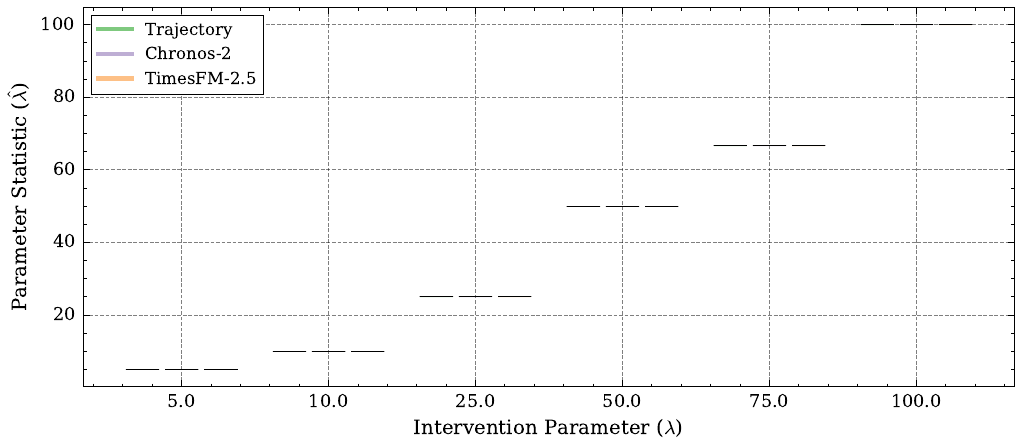}
    \caption{Distributions of parameter statistics for trajectory and model outputs across noise realizations in Experiment 3 ($n=50$). Both models' distributions align with the trajectory distribution and display no variance.}
    \label{fig:exp3_box}
\end{figure}

\begin{figure}[htbp]
    \centering
    \includegraphics[width=1\linewidth]{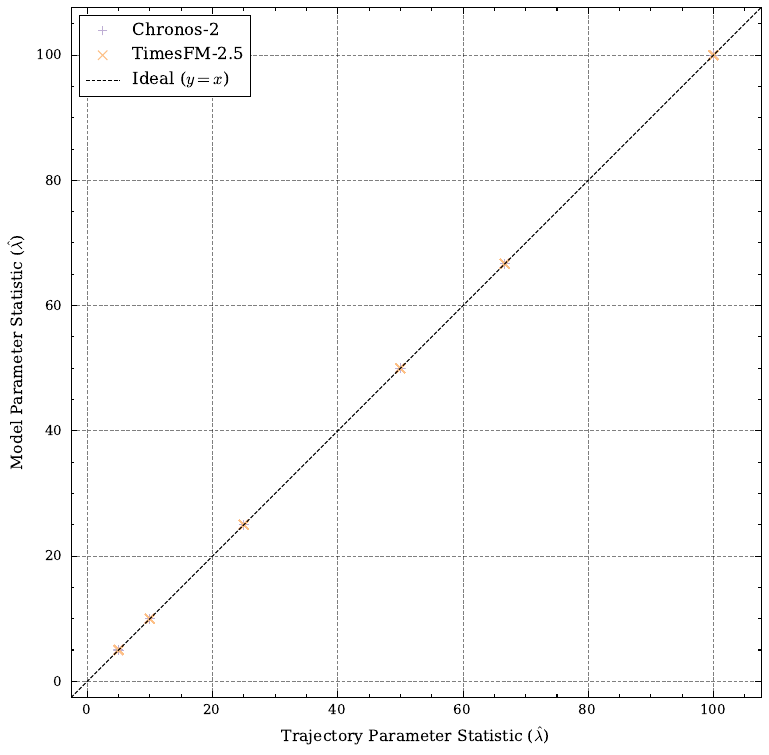}
    \caption{Comparison of model parameter statistic $\delta(M(\mathbf{y}))$ against trajectory parameter statistic $\delta(\mathbf{y})$ for Experiment 3 ($n = 6 \times 50 = 300$ per model). Both models show ideal alignment with no visible deviations.}
    \label{fig:exp3_scatter}
\end{figure}

In \textbf{Experiment 3}, both models preserve the wavelength of the harmonic oscillator across the full range of interventions as shown in Figures \ref{fig:exp3_traj}, \ref{fig:exp3_box} and \ref{fig:exp3_scatter}. For our intervention range, we did not find a value for which any of two the models' estimated wavelength parameter statistic $\delta(M(\mathbf{y}))$ does not align with the intervention parameter $\lambda$ and the trajectory parameter statistic $\delta(\mathbf{y})$, with $\delta$ being $\hat{\lambda}$. Figure \ref{fig:exp3_traj} also shows that some amplitude values are not exactly preserved, as they show small variations along interventions.

\begin{figure}[htbp]
    \centering
    \includegraphics[width=1\linewidth]{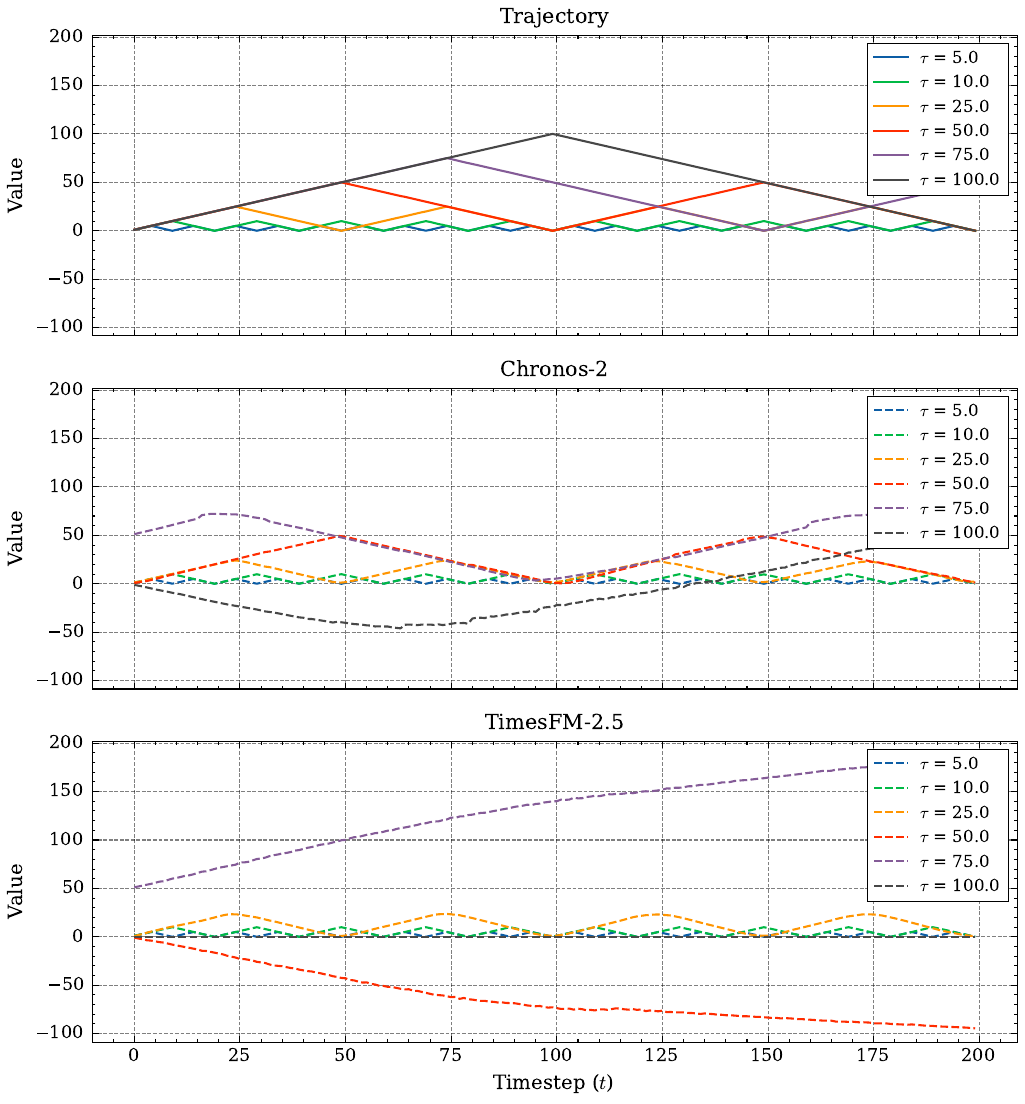}
    \caption{Trajectories across intervention sweep for a single noise realization in Experiment 4. \texttt{Chronos-2} and \texttt{TimesFM-2.5}  fail to preserve regimes for interventions where $\tau \geq 50$ and $\tau \geq 25$, respectively.}
    \label{fig:exp4_traj}
\end{figure}

\begin{figure}[htbp]
    \centering
    \includegraphics[width=1\linewidth]{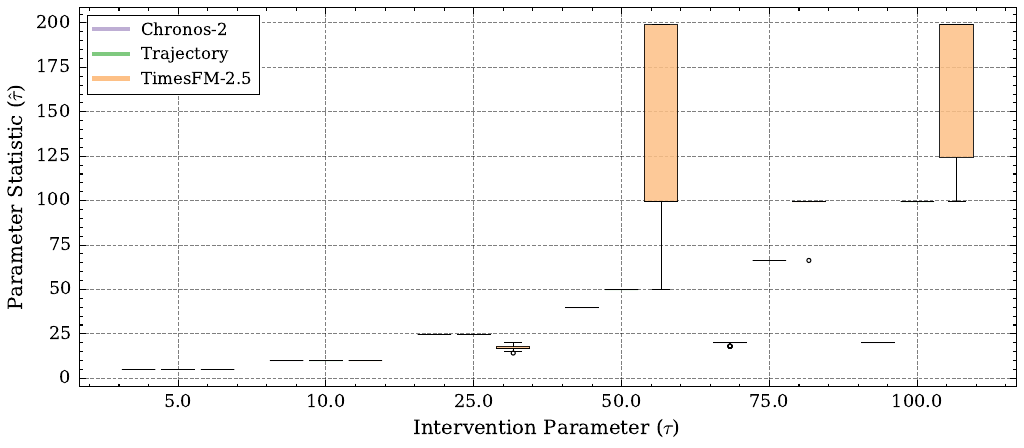}
    \caption{Distributions of parameter statistics for trajectory and model outputs across noise realizations in Experiment 4 ($n=50$). \texttt{TimesFM-2.5} exhibits stronger variance and deviation from trajectory parameter statistic than \texttt{Chronos-2} for higher values of $\tau$. Further \texttt{TimesFM-2.5} seems to overestimate $\tau$ while \texttt{Chronos-2} seems to underestimate it with increasing values of $\tau$.}
    \label{fig:exp4_box}
\end{figure}

\begin{figure}[htbp]
    \centering
    \includegraphics[width=1\linewidth]{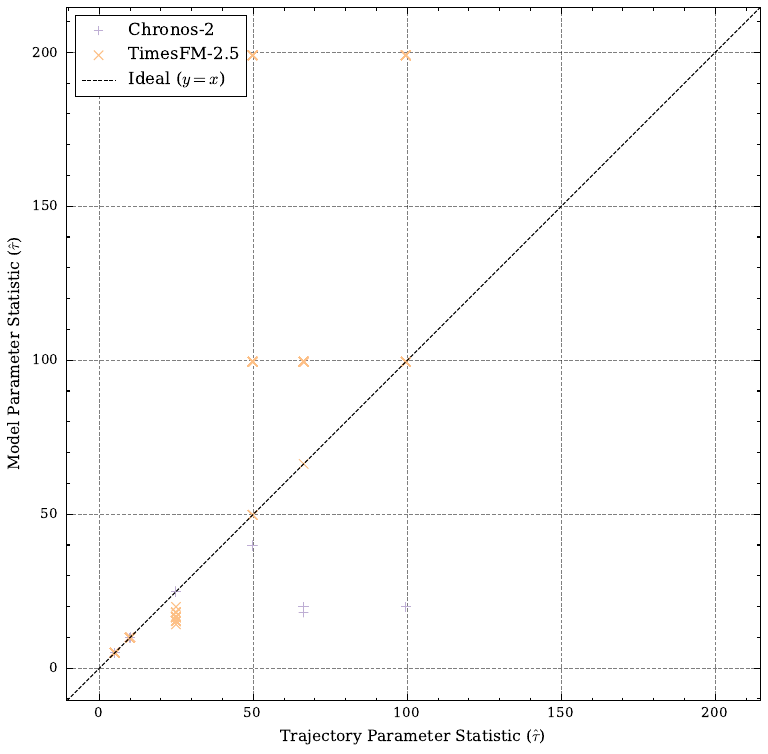}
    \caption{Comparison of model parameter statistic $\delta(M(\mathbf{y}))$ against trajectory parameter statistic $\delta(\mathbf{y})$ for Experiment 4 ($n = 6 \times 50 = 300$ per model). \texttt{TimesFM-2.5} seems to produce outputs with higher $\hat{\tau}$ than the trajectories for $\tau \geq 50$, while \texttt{Chronos-2} produces outputs with lower $\hat{\tau}$ for $\tau \geq 50$.}
    \label{fig:exp4_scatter}
\end{figure}

\textbf{Experiment 4} indicates failure modes for both models in preserving the regime switch pattern for some interventions. Figure \ref{fig:exp4_traj} shows that \texttt{Chronos-2} and \texttt{TimesFM-2.5}  fail to preserve regimes for interventions where $\tau \geq 50$ and $\tau \geq 25$, respectively. Figure \ref{fig:exp4_scatter} shows that while \texttt{TimesFM-2.5} maintains some noise realizations on the ideal line for a given intervention, \texttt{Chronos-2} seems to deviate less. Figure \ref{fig:exp4_box} also seems to confirm these insights.

\begin{figure}[htbp]
    \centering
    \includegraphics[width=1\linewidth]{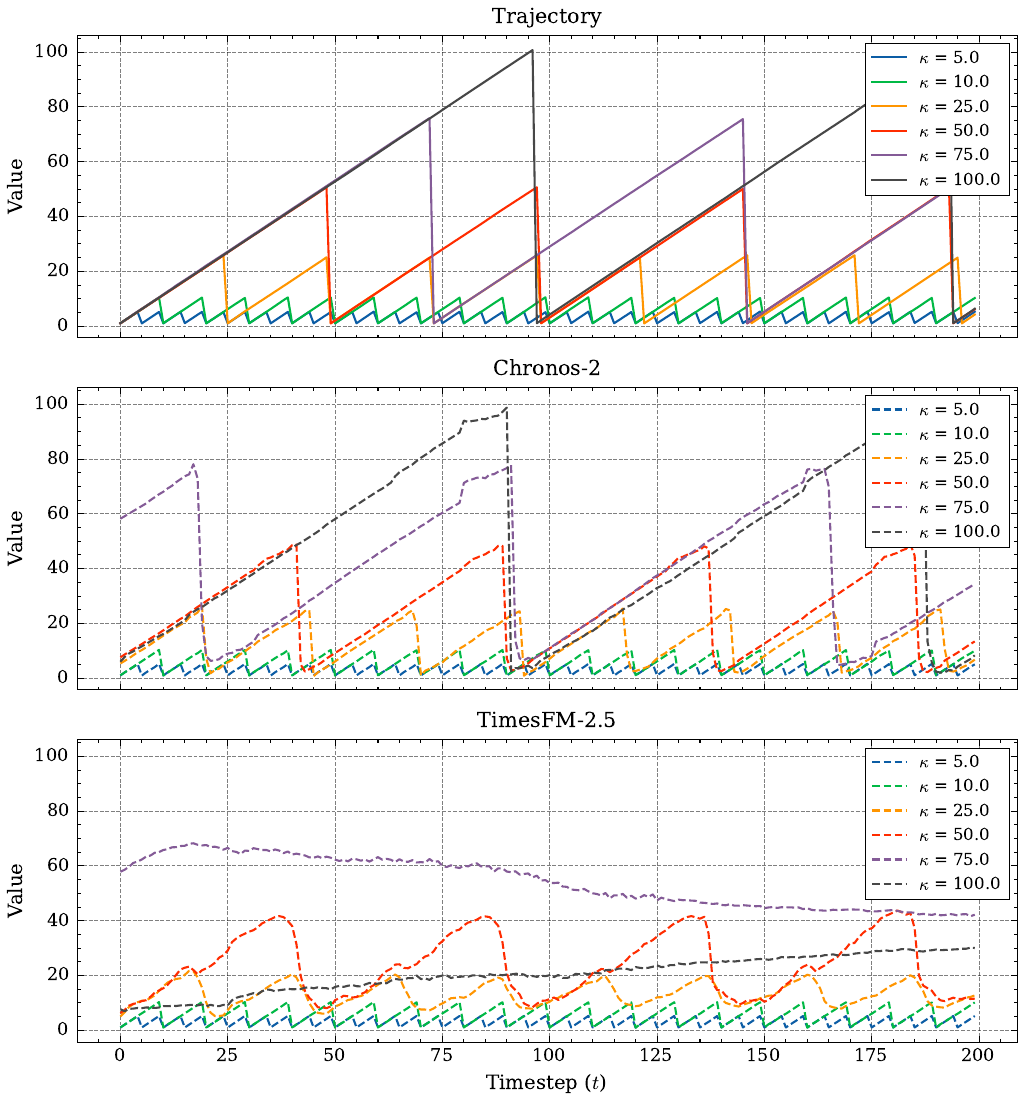}
    \caption{Trajectories across intervention sweep for a single noise realization in Experiment 5. \texttt{Chronos-2} preserves the threshold pattern across the full range of interventions while \texttt{TimesFM-2.5} seems to start smoothing at $\kappa \geq 25$ and  loses pattern completely for $\kappa \geq 50$.}
    \label{fig:exp5_traj}
\end{figure}

\begin{figure}[htbp]
    \centering
    \includegraphics[width=1\linewidth]{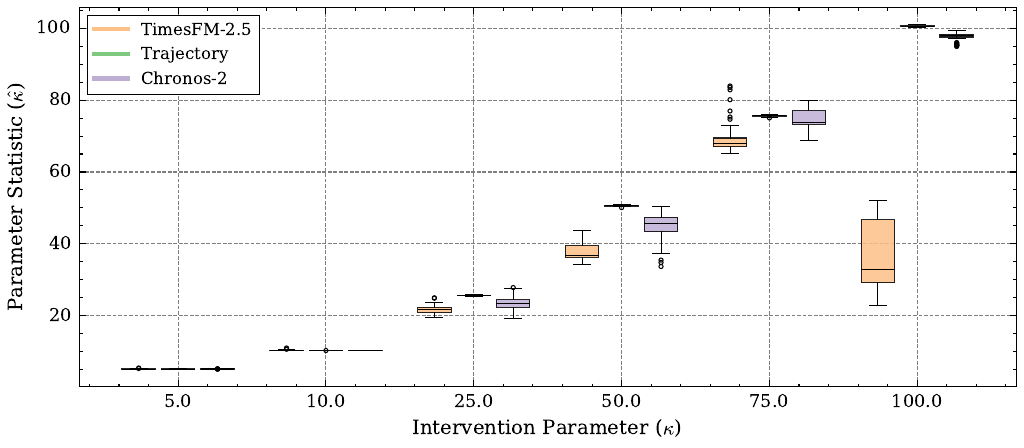}
    \caption{Distributions of parameter statistics for trajectory and model outputs across noise realizations in Experiment 5 ($n=50$). \texttt{Chronos-2} is better at preserving the threshold pattern across the range of interventions than \texttt{TimesFM-2.5}.}
    \label{fig:exp5_box}
\end{figure}

\begin{figure}[htbp]
    \centering
    \includegraphics[width=1\linewidth]{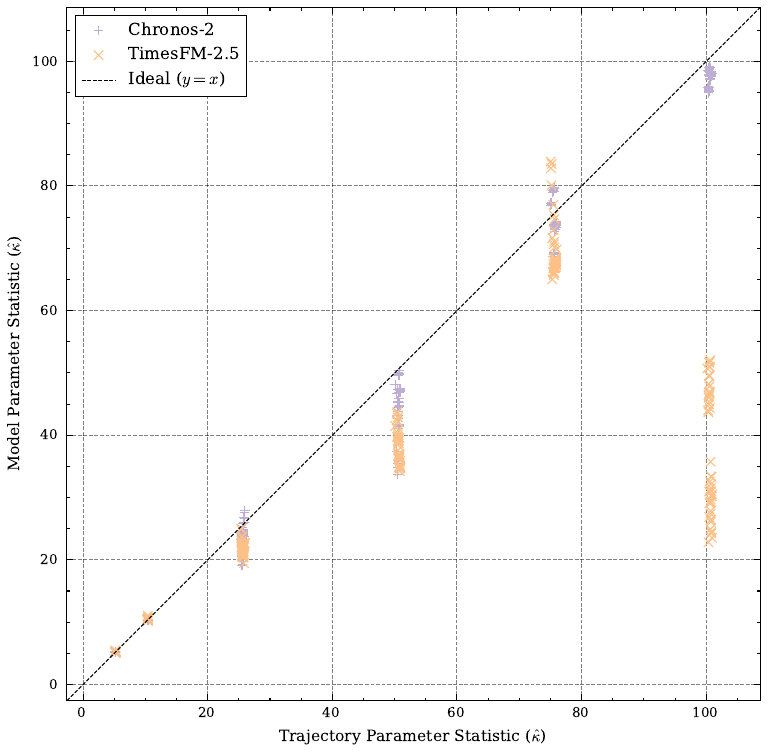}
    \caption{Comparison of model parameter statistic $\delta(M(\mathbf{y}))$ against trajectory parameter statistic $\delta(\mathbf{y})$ for Experiment 5 ($n = 6 \times 50 = 300$ per model). \texttt{Chronos-2} displays better aligment than \texttt{TimesFM-2.5}.}
    \label{fig:exp5_scatter}
\end{figure}

For the energy-release generator in \textbf{Experiment 5}, Figure \ref{fig:exp5_traj} shows that \texttt{Chronos-2} preserves the threshold pattern across the full range of interventions while \texttt{TimesFM-2.5} seems to start smoothing at $\kappa \geq 25$ and  loses pattern completely for $\kappa \geq 50$. Figure \ref{fig:exp5_scatter} shows that for $\hat{\kappa} \approx 5$ and $\hat{\kappa}  \approx 10$ both models' $\delta(M(\mathbf{y}))$ aligns with the trajectory's $\delta(\mathbf{y})$. For $\hat{\kappa} \approx 25$ dispersion begins, while at $\hat{\kappa}  \approx 50$ both models begin to underestimate $\kappa$. With $\hat{\kappa} \approx 75$ both models seem less biased toward underestimation but express higher variance in $\delta(M(\mathbf{y}))$ values and with $\hat{\kappa} \approx 100$, \texttt{TimesFM-2.5} underestimates $\kappa$ noticably more than \texttt{Chronos-2}. Figure \ref{fig:exp5_box} also suggests that \texttt{Chronos-2} is better at preserving the threshold pattern across the range of interventions than \texttt{TimesFM-2.5}.

\begin{figure}[htbp]
    \centering
    \includegraphics[width=1\linewidth]{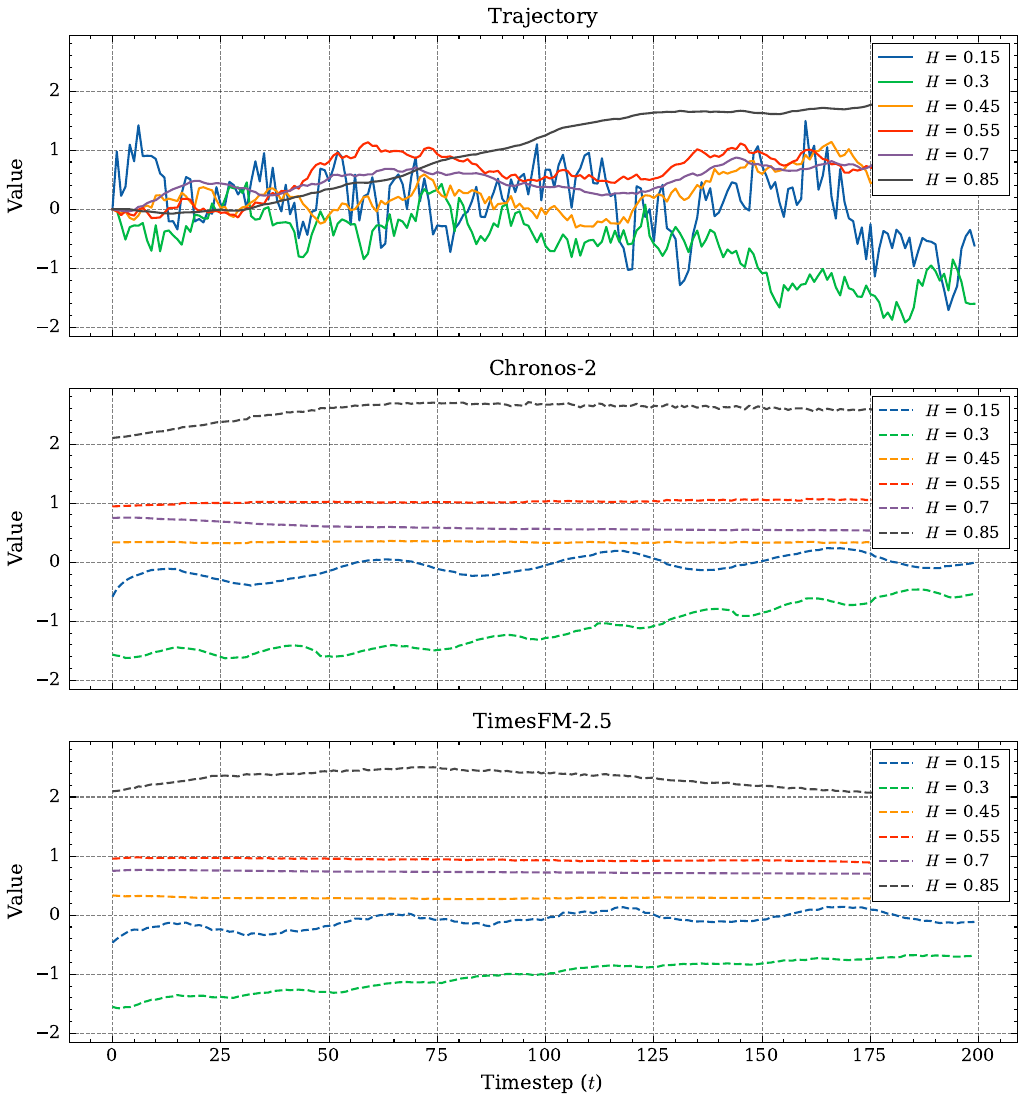}
    \caption{Trajectories across intervention sweep for a single noise realization in Experiment 6. Both models struggle with preserving the jaggedness for interventions where $H < 0.5$ and both smooth out the input trajectory $\mathbf{y}$ across interventions.}
    \label{fig:exp6_traj}
\end{figure}

\begin{figure}[htbp]
    \centering
    \includegraphics[width=1\linewidth]{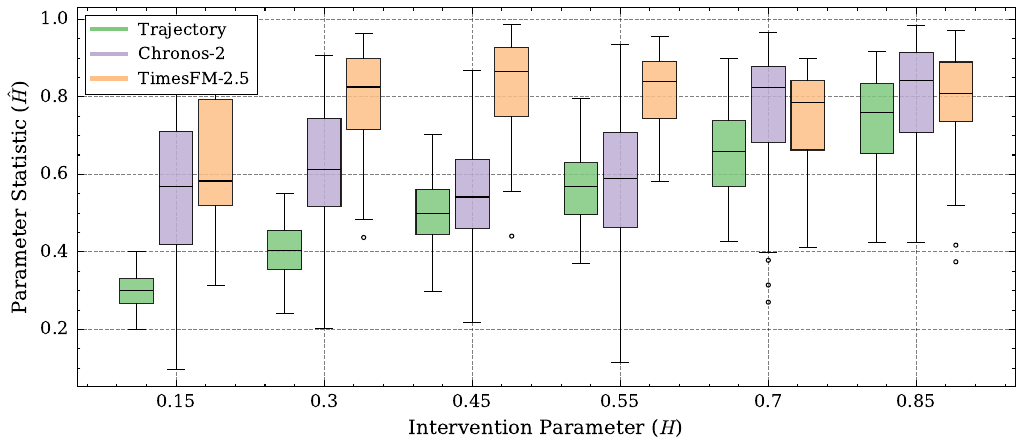}
    \caption{Distributions of parameter statistics for trajectory and model outputs across noise realizations in Experiment 6 ($n=50$). Distributions of both models overestimate $H$ relative to the trajectory distribution, while \texttt{Chronos-2} does so less than \texttt{TimesFM-2.5}.}
    \label{fig:exp6_box}
\end{figure}

\begin{figure}[htbp]
    \centering
    \includegraphics[width=1\linewidth]{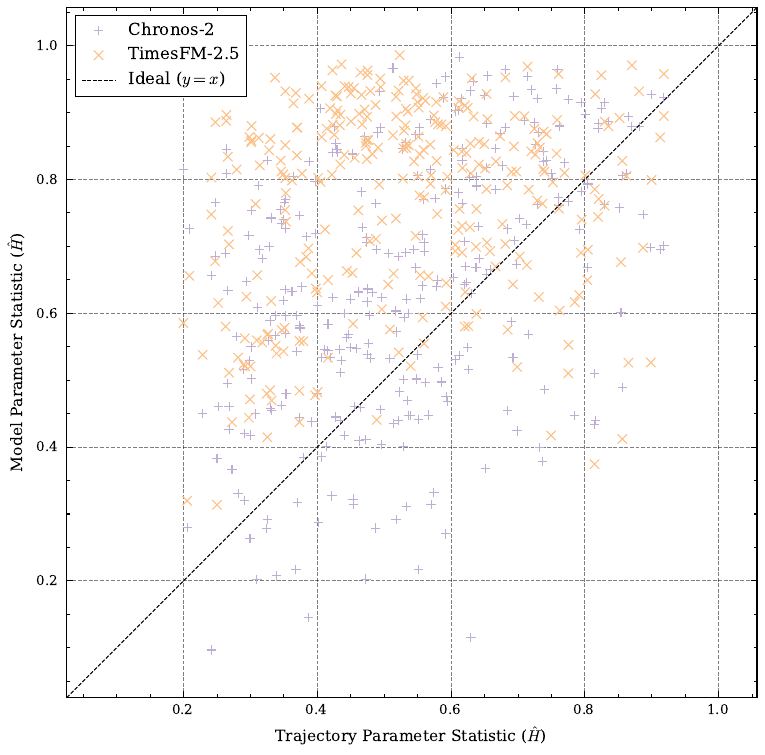}
    \caption{Comparison of model parameter statistic $\delta(M(\mathbf{y}))$ against trajectory parameter statistic $\delta(\mathbf{y})$ for Experiment 6 ($n = 6 \times 50 = 300$ per model). Both models are biased towards overestimating $\hat{H}$.}
    \label{fig:exp6_scatter}
\end{figure}

\textbf{Experiment 6} tests how well each model preservers the Hurst exponent of a fractional Brownian motion generator. The model output $M(\mathbf{y})$ displayed in Figure \ref{fig:exp6_traj} suggests that both models struggle with preserving the jaggedness for interventions where $H < 0.5$ and both smooth out the input trajectory $\mathbf{y}$ across interventions. Figure \ref{fig:exp6_scatter} confirms this, as it shows that both models are biased towards overestimating $\hat{H}$, therefore being biased towards persistence of trends. The figure also reveals that \texttt{Chronos-2} underestimates $\hat{H}$ more for lower values of $H$ than \texttt{TimesFM-2.5}. \texttt{TimesFM-2.5} only shows cases of underestimation for higher values of $H$. Figure \ref{fig:exp6_box} is also aligned with these findings, as distributions of both models overestimate $H$ relative to the trajectory distribution, while \texttt{Chronos-2} does so less than \texttt{TimesFM-2.5}.

In summary, through the application of our causal analysis framework for time series foundation models, we were able to identify stable configurations, as well several biases and failure modes across our experiments. We find that \texttt{Chronos-2} and \texttt{TimesFM-2.5} preserved generator characteristics for a random walk with drift and a harmonic oscillator, within the experimental configurations. We further identify a potential bias towards overestimating persistence in time series in the form of autocorrelation or long-term dependence for both models. We also demonstrate that both models fail to preserve regime switch pattern within our range of interventions and that \texttt{TimesFM-2.5} deviates strongly from the threshold pattern in Experiment 5 for higher threshold values while \texttt{Chronos-2} only deviates weakly. In the next section, we compare our findings to previous work, consider explanations for our findings and discuss implications of our findings for model development and application.

\section{Discussion}
\label{discussion}

With the findings from the previous section, several questions arise. First, do we see our findings reflected in benchmark evaluations of \texttt{Chronos-2} and \texttt{TimesFM-2.5}? Second, do we find evidence in the training and architecture that could explain our findings? And lastly, what are the implications of our findings for time series foundation model development and application?

To address whether our findings are reflected in previous benchmark comparisons, we review reported performance results in \cite{ansariChronos2UniversalForecasting2025} and \cite{dasDecoderOnlyFoundationModel2024}. \cite{ansariChronos2UniversalForecasting2025}, introducing \texttt{Chronos-2}, compares the model against competitor models on \texttt{fevbench}, \texttt{GIFT-Eval} and \texttt{Chronos Benchmark II} for univariate tasks. The study reports \textit{Average Win Rate} and \textit{Skill Score} per model as aggregate metrics for each benchmark without detailing performance on individual time series used in the respective benchmark. This limits our ability to draw direct comparisons between our findings and the benchmark results reported. \cite{ansariChronos2UniversalForecasting2025} also apply \texttt{Chronos-2} to two case studies. First, the Rossmann sales forecasting case study shows a strong smoothing of jagged patterns for the univariate forecasting, in line with our findings in Experiment 1, 2 and 6. Second, for the univariate energy price forecasting case study, the reported forecast aligns with the somewhat cyclical pattern of the time series. We find this in line with our results in Experiment 3. For both case studies, \cite{ansariChronos2UniversalForecasting2025} only report one sample and no quantitative evaluation, therefore we should view our comparison as showing an absence of contradictory findings rather than a proof for generalizability of our \textit{in vitro} findings to the \textit{in vivo} case studies in \cite{ansariChronos2UniversalForecasting2025}.

\cite{dasDecoderOnlyFoundationModel2024} list performance for all datasets used in both the \texttt{Darts} and \texttt{Monash} benchmarks for \texttt{TimesFM-2.5} and competitor models. For the \texttt{Darts} benchmark, \texttt{TimesFM-2.5} shows no clear domination while it does so for \texttt{Monash} with mixed rankings across datasets in both benchmarks. We do not find a clear domination for the listed datasets where we might assume trend or cyclical patterns neither in the \texttt{Darts} nor the \texttt{Monash} benchmark. Even if we are confident that a dataset exhibits a certain time series pattern, the comparison along competitor models and not along datasets does not allow us to assess whether findings in our experiments transfer to real-world datasets. For example, we would expect that \texttt{TimesFM-2.5} shows lower scale-adjusted error on a dataset with cyclical pattern or drift than for one with regime switches or a threshold pattern. Our review of both studies therefore suggest that their findings do not seem to contradict our experimental results, yet we also find limited comparability. Preferably, we would select real-world datasets from data-generating processes with assumed patterns and validate our \textit{in vitro} findings against these.

To answer our second question, whether we find evidence explaining our results, we inspect the training data used for both time series foundation models. \texttt{Chronos-2} is trained on 23 different datasets, some of them synthetically generated \citep{ansariChronos2UniversalForecasting2025}. Five of the datasets are under the category of energy and six under transportation. If we assume that these datasets possess cyclical patterns due physical processes such as day-night cycles or workdays and weekend, then almost half of the datasets represent this time series pattern. This emphasis on cyclical patterns in the training data would align with our findings of preservation of wavelength in Experiment 3. To create the univariate synthetic dataset, \cite{ansariChronos2UniversalForecasting2025} uses \texttt{TSI} to create combinations of trend, seasonality and irregularity. We find that this is in line with our results for Experiments 1 and 3 in which \texttt{Chronos-2} demonstrated preservation of drift in a random walk and wavelength in a harmonic oscillator across the full range of interventions. For multivariate synthetic data, \cite{ansariChronos2UniversalForecasting2025} report using autoregressive models, exponential smoothing models, \texttt{TSI} and \texttt{KernelSynth}. Taken together, the training data might be biased towards trend and cyclical patterns which would explain our findings in Experiment 1 and 3.

\cite{dasDecoderOnlyFoundationModel2024} list four out of the nineteen training datasets as trend. The categories eletricity, traffic and weather are represented by one dataset each and plausibly cyclical patterns. \cite{dasDecoderOnlyFoundationModel2024} also mention that majority of their datasets exhibit a periodic pattern. They further report that the synthetic data consists of piece-wise linear trends, ARMA processes and seasonal patterns. The provided illustrative examples also heavily lean towards cyclical time series. Similar to \cite{ansariChronos2UniversalForecasting2025}, we also find evidence in the training data that indicates a bias towards trend and cyclical patterns. The therefore lesser representation of other time series patterns could explain the observed failure modes in Experiment 4 and 5 as well as the bias towards overestimating persistence for lower persistence values of $\beta$ and $H$, suggested by the findings of Experiment 2 and 6.

Having discussed our findings within the context of the original works, we now may concern ourselves with the implications of these findings. We can group them into (i) implications for future model development and (ii) implications for model selection for applications. For model development (i), our empirical findings and review of original works indicate that more diverse training data might be beneficial, specifically targeting the identified failure modes and biases. To what extent more diverse training data is able to remedy those might only be found out through experimentation, as model architecture, other factors, or a combination of these factors might be the actual cause. \cite{ansariChronos2UniversalForecasting2025} also show that a version of \texttt{Chronos-2} that was only trained on synthetic data performed close to the version trained with real and synthetic datasets and suggest that synthetic-only training might be a viable pretraining strategy for time series foundation models. Building on this evidence, we might replace real-world training datasets that have fixed empiricial distributions with parameterized synthetic generators that allow us to pretrain time series foundation models across a wide range of generator configurations, ensuring safe-use boundaries. For model selection (ii), our findings also indicate differences between \texttt{Chronos-2} and \texttt{TimesFM-2.5} for different applications. While the discussed benchmark evaluations give comparisons of competitor models across different datasets, our analysis demonstrates how a given model performs across a range of potential realizations of a given time series pattern. This additional insight might be particularly relevant for practitioners with domain knowledge of their application, allowing them to specify the kind of time series patterns they expect in their application. In this situation understanding how time series foundation models perform against a range of pattern realizations might be more insightful for model selection than how the models rank across a range of different datasets exhibiting a diverse range of, partially unknown, time series patterns. If we want to predict foot traffic and assume a cyclical day and night pattern as well as weekday and weekend pattern and need to choose between \texttt{Chronos-2} and \texttt{TimesFM-2.5}, we might be more interested in which model preserves cyclical structure such as wavelength rather than understanding how both models perform across a wide array of datasets from different applications. We summarize our recommendations for model selection by time series pattern in Table \ref{tab:model_comparison_experiments}.

\begin{table}[htbp]
\centering
\caption{Comparison of model behaviors, failure modes, and model selection recommendations across patterns.}
\label{tab:model_comparison_experiments}
\footnotesize
\setlength{\tabcolsep}{2pt} 
\begin{tabularx}{\linewidth}{X X X c}
\toprule
\textbf{Pattern} & \textbf{\texttt{Chronos-2} Behavior} & \textbf{\texttt{TimesFM-2.5} Behavior} & \textbf{Recommendation} \\
\midrule
Random Walk & Preserves drift magnitude ($\hat{\mu}$ closer to trajectory) & Underestimates drift magnitude (bias toward zero) & \texttt{Chronos-2}\\
\addlinespace
AR(1) & Output collapses to line & Output collapses to line & Avoid \\
\addlinespace
Harmonic Oscillator & Preserves $\hat{\lambda}$; cuts amplitude for some wavelengths & Preserves $\hat{\lambda}$; cuts amplitude for some wavelengths & Tie \\
\addlinespace
Regime Switch & Preserves regimes and dwell time up to $\tau = 50$ & Deviates earlier; leaves ideal line at $\tau = 25$ & \texttt{Chronos-2} \\
\addlinespace
Energy-release & Preserves pattern across range; underestimates at $\kappa = 100$ & Smoothes at $25 < \kappa < 50$; strong underestimation at $\kappa = 100$ & \texttt{Chronos-2} \\
\addlinespace
fBm & Overestimates $\hat{H}$; strong smoothing for $H < 0.5$ & Overestimates $\hat{H}$; strong smoothing for $H < 0.5$ & Avoid \\
\bottomrule
\end{tabularx}
\end{table}

\section{Conclusion}
\label{conclusion}

We conclude this study by summarizing our four contributions, highlighting limitations and suggesting further research directions. To the best of our knowledge this is the first study that formalizes a causal analysis framework for time series foundation models. We identified dose-response relationships for \texttt{Chronos-2} and \texttt{TimesFM-2.5} across six generators and five interventions each. We identify a potential biases towards overestimating persistence in Experiment 2 and 6 for both models and showcase idiosyncratic failure modes for each of them. Lastly we make concrete recommendations for application-specific model selection based on our experimental results.

Our findings are limited to the specific experimental configurations and models analyzed. We used  a fixed set of generators and intervention parameter per generator, sweeped a limited range of values, and used only one noise scale per experiment. We further only evaluated the preservation of parameter statistics on a fixed window of 200 time steps and used nominal parameter statistic values and did not relate them to the window length. For example, it is not clear whether the failure mode in Experiment 4 is caused by the nominal dwell time or the number of observable regime switches within the input trajectory.

While the limitations mentioned outline the epistemic boundaries of this study, they also signal opportunities for further research. Future work could extend experiments to more generators and time series foundation models. We could create domain specific experiments to analyse model suitability for domains such as climate, finance or sales. We could also adapt the causal analysis framework towards analyzing relationships between structure and behavior, to understand how different components of a time series foundation model affect its behavior in inference. Future model development efforts could focus on mitigating the biases and failure modes we identified while other research efforts could be directed toward \textit{in vivo} validation of our findings through real-world datasets. Together, the structured \textit{in vitro} causal analysis of time series foundation models might improve their capabilities as wells as inform their application and regulation.

\section*{Code and Data Availability}

The codebase for this study can be found at \url{https://github.com/MSCA-DN-Digital-Finance/tsfm_causal_analysis}.

The dataset containing the experimental results can be found at \url{https://zenodo.org/records/22082090}

\section*{Author Contributions}

\textbf{Mathis Jander}: Conceptualization, Methodology, Software, Investigation, Formal Analysis, Data Curation, Writing -- Original Draft, Visualization. \\
\textbf{Wouter van Heeswijk}: Conceptualization, Funding Acquisition, Project Administration, Supervision, Writing -- Review \& Editing.\\
\textbf{Martijn Mes}: Conceptualization, Supervision, Writing -- Review \& Editing.

\section*{Acknowledgments}
The views expressed in this work are those of the authors and do not necessarily reflect those of the European Central Bank or the Eurosystem.

Funded by the European Union. Views and opinions expressed are however those of the author(s) only and do not necessarily reflect those of the European Union or European Research Executive Agency (REA). Neither the European Union nor the granting authority can be held responsible for them.

\noindent
This project has received funding from the Horizon Europe research and innovation programme under the Marie Skłodowska-Curie Grant Agreement No. 101119635

\begin{figure}[htbp]
    \centering
    \includegraphics[width=\linewidth]{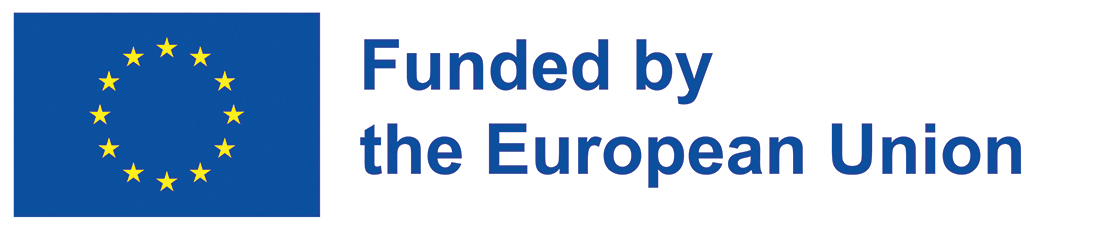}
\end{figure}

\bibliographystyle{apalike}
\bibliography{project} 

\end{document}